\documentclass[letterpaper, 10 pt, conference]{ieeeconf}  

\IEEEoverridecommandlockouts                              

\usepackage{graphicx}
\usepackage{amsmath}
\usepackage{amssymb}
\usepackage{booktabs}
\usepackage{hyperref}
\usepackage{multirow}

\title{\LARGE \bf
Sim-to-Real Domain Adaptation for Lane Detection and Classification in Autonomous Driving}

\author{Chuqing Hu, Sinclair Hudson, Martin Ethier,  \\ Mohammad Al-Sharman, Derek Rayside, and William Melek
\thanks{C. Hu, M. Ethier and W. Melek are with the Department of Mechanical and Mechatronics engineering, 
        University of Waterloo, ON, Canada,  e-mails: 
        {\tt\small cq2hu@uwaterloo.ca, methier@uwaterloo.ca, william.melek@uwaterloo.ca}.}%
\thanks{S. Hudson is with the Department of Computer Science, University of Waterloo,
        ON, Canada,  e-mail:
        {\tt\small sshudson@uwaterloo.ca}}%
\thanks{M. Al-Sharman, D. Rayside are with the Department of Electrical and Computer Engineering, University of Waterloo, ON, Canada, e-mails:
        {\tt\small mkalsharman@uwaterloo.ca, drayside@uwaterloo.ca}. All authors are with WATonomous (University of Waterloo autonomous vehicle team in the SAE AutoDrive Challenge. (watonomous.ca)).}%
}

\begin{document}

\maketitle
\thispagestyle{empty}
\pagestyle{empty}

\begin{abstract}

 While supervised detection and classification frameworks in autonomous driving require large labelled datasets to converge, Unsupervised Domain Adaptation (UDA) approaches, facilitated by synthetic data generated from photo-real simulated environments, are considered low-cost and less time-consuming solutions. In this paper, we propose UDA schemes using adversarial discriminative and generative methods for lane detection and classification applications in autonomous driving. We also present \emph{Simulanes} dataset generator to create a synthetic dataset that is naturalistic utilizing CARLA's vast traffic scenarios and weather conditions. The proposed UDA frameworks take the synthesized dataset with labels as the source domain, whereas the target domain is the unlabelled real-world data. Using adversarial generative and feature discriminators, the learnt models are tuned to predict the lane location and class in the target domain. The proposed techniques are evaluated using both real-world and our synthetic datasets. The results manifest that the proposed methods have shown superiority over other baseline schemes in terms of detection and classification accuracy and consistency. The ablation study reveals that the size of the simulation dataset plays important roles in the classification performance of the proposed methods. Our UDA frameworks are available at \href{https://github.com/anita-hu/sim2real-lane-detection}{github.com/anita-hu/sim2real-lane-detection} and our dataset generator is released at \href{https://github.com/anita-hu/simulanes}{github.com/anita-hu/simulanes}.          

\end{abstract}

\section{INTRODUCTION}

As simulators become increasingly photorealistic and simulation platforms become highly flexible in terms of sensor configuration and environmental setting, synthetic data has growing potential in filling gaps within existing real-world datasets \cite{Synth_foggy_scene_IV21, DA_semseg_IV19}. Nevertheless, there exists a domain difference between the appearance of simulation and real-world data \cite{SimGAN_2017} where domain adaptation techniques can be employed to minimize this difference. While domain adaptation is an extensively researched topic in digit recognition \cite{UNIT_NIPS2017, DRCN_ECCV_2016}, object classification \cite{Deep_CORAL_2016, HoMM_2020} and in recent years sim-to-real object detection \cite{RetinaGAN_ICRA21} and semantic segmentation \cite{ADA_IROS_2017, DA_semseg_IV19}, few works \cite{SynDataAug_2020_CVPR_Workshops, GM_ACCV_2020} explore potential utilization in lane detection for autonomous driving applications.

Lane detection is crucial in autonomous driving systems since it serves as the foundation for path planning decisions and is utilized for vehicle localization in high resolution maps. Lane detection is also widely used in Advanced Driver Assistance Systems (ADAS) such as lane keeping assist and adaptive cruise control \cite{lane_detection_survey}. The state-of-the-art works in lane detection \cite{Liu_2021_ICCV, LaneAF_2021} focus on evaluating against existing open-source datasets. However, training a robust neural network requires a large dataset with labelled data covering a wide range of scenarios and environment conditions. Gathering such dataset, however, is considered costly and time consuming. For tasks such as lane detection and classification, the availability of such datasets can be a limiting factor. Hence, Unsupervised Domain Adaptation (UDA) methods can be utilized in this simulation. 

UDA techniques are designed to transfer knowledge from a labelled source domain to a similar target domain that is fully unlabelled by addressing the domain shift between the two domains \cite{DUDA_2019}. Motivated by the recent work by \cite{GM_ACCV_2020, SS_DUDA_2020} in unsupervised and semi-supervised domain adaptation for learning tile-based lane representations in a top view image, our work explores domain adaptation techniques applied in end-to-end lane detection. 

To explore domain adaptation techniques for use in lane detection, we introduce \emph{Simulanes}, a synthetic data generator for lane detection and classification, which is capable of generating photo-realistic traffic scenarios under a variety of weather conditions. Built upon CARLA \cite{CARLA_2017}, which uses Unreal Engine, \emph{Simulanes} addresses the limitation in photorealism of generated synthetic data from \cite{3DLaneNet_ICCV_2019} used in \cite{GM_ACCV_2020}. We use a synthetic dataset generated using \emph{Simulanes} to evaluate lane detection methods, using it as our source for simulated images of road scenes, with labelled lane types and positions.

In \cite{Wayve_2018}, the domain-invariant latent space learnt in UNIT \cite{UNIT_NIPS2017} through image translation and reconstruction, is exploited to transfer an end-to-end driving policy from a simulation domain to an unlabelled real-world domain. Inspired by their work and recent successes in adversarial methods for domain adaptation \cite{SS_DUDA_2020}, in this paper, we focus on developing unsupervised domain adaptation techniques using adversarial generative and adversarial discriminative approaches for lane detection and classification purposes in autonomous driving. Our proposed method performs end-to-end lane detection and classification on the domain-invariant latent space in UNIT and MUNIT \cite{MUNIT_ECCV_2018}. To further encourage the latent features to be domain-invariant, we introduce adversarial discriminators to predict the domain of the latent feature. To the best of the authors' knowledge, this paper is the first to apply sim-to-real domain adaption for lane detection and classification tasks. In summary, this paper offers the following contributions:
\begin{itemize}
  \item unsupervised domain adaptation techniques using adversarial discriminative and generative methods are proposed for lane detection and classification enhancement in autonomous driving.
  \item a new synthetic data generator, \emph{Simulanes}, for generating synthetic lane detection and classification data, featuring diverse traffic conditions, weather, and surrounding environments. 
  \item the proposed detection and classification frameworks are evaluated using TuSimple \cite{TuSimple_Dataset} and our synthetic dataset, and an ablation study on the effect of simulation dataset size on model performance is carried out.
\end{itemize}

\section{RELATED WORK}

In this section, we provide background on unsupervised domain adaptation methods, unsupervised image-to-image translation methods, and sim-to-real lane detection methods.
While our method ultimately performs lane detection, it applies UDA during training, leveraging unsupervised image-to-image translation methods to minimize the domain gap between real images of driving scenarios and simulated images of driving scenarios.

\subsection{Unsupervised Domain Adaptation}

While early UDA methods match feature distributions between source and target domains, deep UDA methods focus on learning domain-invariant features \cite{SS_DUDA_2020}. These methods can be grouped into discrepancy-based, adversarial discriminative, adversarial generative, and self-supervision methods or a combination of these approaches. 
Discrepancy-based methods, such as \cite{Deep_CORAL_2016, HoMM_2020}, introduce a loss function to minimize the discrepancy between the prediction and/or activation layers from source and target streams. 
Adversarial discriminative methods achieve a similar goal of minimizing domain shift using an adversarial objective \cite{SS_DUDA_2020}. For instance, authors in \cite{ADA_IROS_2017} use a discriminator to align feature distributions, whereas Ganin \emph{et al.} \cite{Domain_Adv_JMLR_2016} minimizes cross-covariance between both features and class predictions.
Instead of explicitly comparing source and target representations, self-supervision-based methods add auxiliary self-supervised learning task(s) e.g. image rotation prediction in \cite{SSDA_IEEE_2019}, to help close the domain gap.
Adversarial generative methods combine a generative adversarial network (GAN) with discriminator(s) at the image and/or feature level. CycleGAN \cite{CycleGAN_2017} is adopted in \cite{CyCADA_PMLR_2018} for semantic segmentation adaptation by utilizing feature and image discriminators and enforcing cycle consistency. \cite{3CATN_ACM_2019} uses two feature translators with cycle consistency and conditions the adversarial networks with the cross-covariance of learned features and classifier predictions as in \cite{CADA_NEURIPS2018}. 
While discrepancy-based and self-supervision-based methods are easier to optimize, for more challenging tasks like object detection and semantic segmentation, adversarial learning-based methods are more effective due to their strength in local feature alignment \cite{SS_DUDA_2020}. 

\subsection{Unsupervised Image-to-Image Translation}

Unsupervised image-to-image methods such as \cite{UNIT_NIPS2017, MUNIT_ECCV_2018, DRIT_ECCV_2018} learn a shared latent space in order to translate an image from one domain to another. \cite{UNIT_NIPS2017} uses variational autoencoders (VAEs) to map images to the shared latent space while GANs generate corresponding images in two domains. To increase the diversity of generated images, \cite{MUNIT_ECCV_2018, DRIT_ECCV_2018} breaks down an image representation into a domain-invariant content code and domain-specific style code. \cite{UNIT_NIPS2017} and \cite{DRIT_ECCV_2018} explore different approaches when applying their method for domain adaption. \cite{UNIT_NIPS2017} employs a multi-task learning framework where the network learns image translation and classification using high level features in the discriminator. \cite{DRIT_ECCV_2018} takes a two-stage approach by first training an image translation network from source to target then training a classifier with translated images in the labelled source domain. 

\subsection{Sim-to-Real for Lane Detection}

Recent works, \cite{SynDataAug_2020_CVPR_Workshops} and \cite{GM_ACCV_2020}, investigate domain adaption for lane detection, leveraging synthetic data to enhance the available real-world data.
\cite{SynDataAug_2020_CVPR_Workshops} improves real-world performance in autonomous driving perception tasks including lane detection by mixing GAN translated simulation images with labelled real-world data during training. 
Garnett \emph{et al.} \cite{GM_ACCV_2020} is the first to apply unsupervised and semi-supervised domain adaptation for lane detection using a two stage approach similar to \cite{DRIT_ECCV_2018} with CycleGAN \cite{CycleGAN_2017}. In the second stage, \cite{GM_ACCV_2020} trains lane detection in conjunction with lane segmentation and self-supervised view orientation prediction tasks. 
\cite{GM_ACCV_2020} adopts a segment-based approach to lane detection which requires clustering on the model output for lane-based evaluation. By contrast, our work explores an end-to-end approach to lane detection using row anchors as in \cite{CULane_Dataset, UFLD_ECCV_2020}. \cite{GM_ACCV_2020} uses the method in \cite{3DLaneNet_ICCV_2019} to generate scenes with randomized lane topology, road textures and objects within Blender as labelled training data. While this approach is capable of generating scenes with high variability in lane topology and 3D geometry, it has limited variety in the appearance of lanes and the surrounding environment. We aim to fill this gap with our simulation framework using CARLA \cite{CARLA_2017} to generate photo-realistic scenes within urban, rural and highway environments.

\section{METHOD}

The challenge of utilizing simulation data to develop a model for a task without real-world labels can be formulated as an UDA problem. Here, the source domain is a dataset generated from simulation $X_{sim}$ with labels $Y_{sim}$ and the target domain is a real-world dataset $X_{real}$ without any labels. Naturally, the images from simulation $X_{sim}$ and the images from the real-world dataset $X_{real}$ are unpaired. The goal of the learned model is to correctly predict the labels of the target domain $Y_{real}$. 

\subsection{Simulanes: Simulation Data Generator}

Real-world driving is heterogeneous with diverse traffic conditions, weather, and surrounding environment. Thus, the diversity in simulated scenarios is crucial for the model to adapt well in the real-world. There are many open source simulators for autonomous driving, namely CARLA and LGSVL \cite{LGSVL_2020} as the state-of-the-art for end to end testing with high quality simulation environments \cite{SimulatorSurvey_2021}. In this paper, we chose CARLA for generating our simulation dataset due to its rich content of premade maps covering urban, rural and highway scenarios, in addition to its flexible Python API.

\begin{figure}[t]
\centering
\includegraphics[width=0.95\columnwidth]{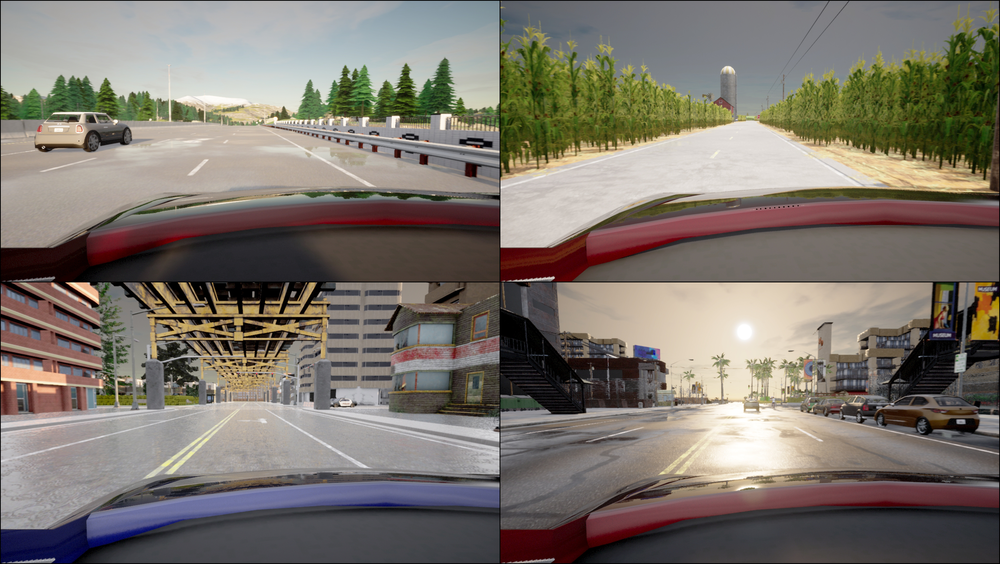} 
\caption{Sample images from \emph{Simulanes} showing highway, rural and urban scenarios with varied weather.}
\label{fig1}
\end{figure}

Our simulation data generator, \emph{Simulanes}, generates various simulation scenarios in urban, rural and highway environments with 15 lane classes and dynamic weather. Figure \ref{fig1} shows samples from our synthetic dataset. Pedestrian and vehicle actors are randomly generated and placed to roam on the maps, increasing the difficulty of the dataset through occlusions. Following TuSimple \cite{TuSimple_Dataset} and CULane \cite{CULane_Dataset}, we limit the maximum number of lanes to 4 adjacent to the vehicle and use row anchors for the labels. Since lane location labels are not directly given by the CARLA simulator, we generated the labels by utilizing CARLA's waypoint system. Waypoints in CARLA are predefined locations for the vehicle autopilot to follow and are located in the center of the lanes. To get lane location labels, the waypoints in the current lane were shifted right and left by $W/2$, where $W$ is the width of the lane given by the simulator. These shifted waypoints were then projected to the camera frame and a spline was fit to them in order to generate labels along the predefined row anchors. The class labels are given by the simulator, and can be one of 15 different classes. To generate a dataset with $N$ frames, we divide $N$ evenly across all available maps. From the default CARLA maps, towns 1, 3, 4, 5, 7, and 10 were used, while towns 2 and 6 were not used due to discrepancies between the extracted lane location labels and the lane locations in the image. For each map, the vehicle actor is spawned at a random location and would roam randomly. Dynamic weather is achieved through changing the sun's position smoothly with time as a sinusoidal function and generating storms occasionally which affects the look of the environment through variables such as cloudiness, precipitation and precipitation deposits. To avoid saving multiple frames at the same location, we check that the vehicle has moved from the previous frame location and respawn the vehicle if it has stopped for too long. 

\subsection{Lane Detection Model}

As we apply the proposed sim-to-real algorithm for lane detection, we adopt an end-to-end approach and use Ultra-Fast-Lane-Detection (UFLD) \cite{UFLD_ECCV_2020} as our base network. We chose UFLD due to its light weight architecture achieving 300+ FPS with the same input resolution while having comparable performance to state-of-the-art methods. UFLD formulates the lane detection task as a row-based selecting method where each lane is represented by a series of horizontal positions at predefined rows, i.e., row anchors. For each row anchor, the position is divided into $w$ gridding cells. For the $i$-th lane and $j$-th row anchor, the prediction of position becomes a classification problem where the model outputs the probability, $P_{i,j}$, of selecting $(w + 1)$ gridding cells. The extra dimension in the output indicates the absence of lane. The lane location loss is given by:
\begin{equation}
    L_{loc} = \sum_{i=1}^{C} \sum_{j=1}^{h} L_{CE}(P_{i,j}, T_{i,j})
\end{equation}
where $C$ is the maximum number of lanes, $h$ is the number of row anchors, and $T_{i,j}$ is the one-hot label of the correct position of the $i$-th lane at the $j$-th row anchor. To ensure that the predicted lanes are continuous, similarity loss $L_{sim}$ is added to constrain the distribution of classification vectors over adjacent row anchors. This is done by calculating the L1 norm as follows:
\begin{equation}
    L_{sim} = \sum_{i=1}^{C} \sum_{j=1}^{h-1} \lVert P_{i,j} - P_{i,j+1} \rVert_1
\end{equation}

An auxiliary segmentation branch is proposed in \cite{UFLD_ECCV_2020} to model local features by aggregating features at multi-scales and is only used during training. Following UFLD, cross entropy loss is used for the segmentation loss $L_{seg}$. For lane classification, a small branch with fully connected (FC) layers is added which receives the same features as the FC layers for lane location prediction. Lane classification loss $L_{cls}$ is also using cross entropy loss.

The overall supervised lane detection and classification task loss is formulated as:
\begin{equation}
    L_{task} = L_{loc} + \alpha L_{sim} + \beta L_{seg} + \gamma L_{cls}
\end{equation}
where $\alpha$, $\beta$ and $\gamma$ are loss coefficients.

\begin{figure*}
    \centering
    \includegraphics[width=0.95\textwidth]{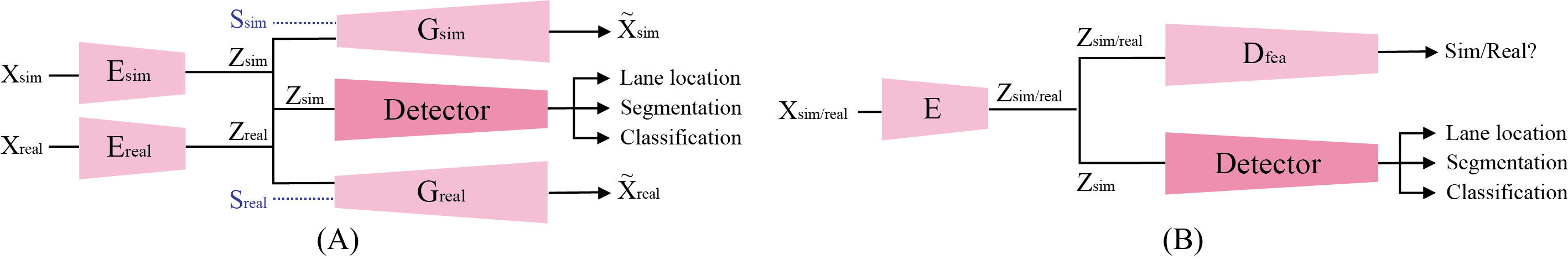}
    \caption{Proposed adversarial generative (A) and adversarial discriminative methods (B). Both UNIT and MUNIT are represented in (A) with generator inputs shown for image translation. MUNIT's additional style input is shown in blue dotted lines. For simplicity, MUNIT's style encoder output is omitted as it is not used in image translation.}
    \label{models}
\end{figure*}

\subsection{Adversarial Generative and Descriminative Models}

To mitigate domain shift in the UDA setting, we employ an adversarial generative approach using UNIT and MUNIT, and an adversarial discriminative approach using a feature discriminator. Our proposed architectures are presented in Figure \ref{models}.

\paragraph{Adversarial generative} The UNIT framework utilizes an encoder-generator pair $\{E_d, G_d\}$ for each image domain where $d \in [sim, real]$. The encoder $E_d$ is a VAE that maps an image $X_d$ to a latent space $\mathbb{Z}$. The generator $G_d$ then reconstructs the input image from a random-perturbed version of the latent code $Z_d$. Image translation or reconstruction is implemented as follows:
\begin{equation}
    Z_d = E_d(X_d) + \eta
\end{equation}
\begin{equation}
    \tilde{X_d} = G_d(Z_d)
\end{equation}
where $\eta$ is a random noise sampled from $\mathcal{N}(0, 1)$ and is only added during training. Image translation is where $E_d$ and $G_d$ are from different domains whereas image reconstruction is when they are from the same domain. 

MUNIT is similar to UNIT where an encoder-generator pair is used for each domain. However, the encoder $E_d$ consists of a context encoder and a style encoder. The two encoders map the image to a context code $Z_d$ and a style code $S_d$. The generator $G_d$ reconstructs input images using the context code and style code. Image translation is done using the context code and a random style code from a Gaussian distribution $\mathcal{N}(0, I)$. 

To ensure cyclic consistency, cyclic reconstruction is used where the input is translated to the other domain and then back. Image reconstruction and cyclic reconstruction are learned via $L_{recon}$ and $L_{cyc{\text -}recon}$ which calculates the L1 loss i.e. mean absolute error between the reconstructed image and the original image. 

The encoder-generator pairs are trained along with two domain adversarial discriminators $D_{sim}$ and $D_{real}$. The discriminator $D_d$ learns to predict true for real images in the domain $d$ and false for the generated images by $G_d$. The image translation task optimizes the encoder-generator and discriminator using the Least-Squares Generative Adversarial Network (LSGAN) objective from \cite{LSGAN_ICCV_2017}. Here the generator $G_d$ learns to generate images that resemble real images from domain $d$ to trick the corresponding discriminator $D_d$. This is formulated as:
\begin{equation}
\begin{aligned}
    L_{LSGAN}(D) = & \mathbb{E}_{X \sim p_{data}(X)}[(D(X)-1)^2] + \\
    & \mathbb{E}_{Z \sim p_z(Z)}[(D(G(Z))-0)^2]
\end{aligned}
\end{equation}
\begin{equation}
    L_{LSGAN}(G) = \mathbb{E}_{Z \sim p_z(z)}[(D(G(Z))-1)^2]
\end{equation}
To ensure the translated image contains similar semantics as the original, we compare their features using a pretrained VGG16 network. The original and translated images are passed through the VGG network. The features from the last convolution layer are normalized with Instance Norm and find perceptual loss $L_{VGG}$ which computes the feature difference via Mean Squared Error (MSE) loss.

The lane detection task is learned with domain-invariant latent features $Z_{sim}$ as input. The detector is a Convolutional Neural Network (CNN) that is trained in a supervised manner with $L_{task}$ and $L_{cyc{\text -}task}$ using labelled simulation data. $L_{cyc{\text -}task}$ is the cyclic task loss where $X_{sim}$ is first translated to the real domain then encoded using $E_{real}$ and passed to the detector. For the UNIT framework, the total loss for the encoder-generator pairs and the lane detector is the sum of the following losses weighted by $\lambda_i$:
\begin{equation}
\label{total_loss}
\begin{aligned}
    L_{total} = & \lambda_0 L_{recon} + \lambda_1 L_{cyc{\text -}recon} + \lambda_2 L_{LSGAN}(G) + \\
    & \lambda_3 L_{task} + \lambda_4 L_{cyc{\text -}task} + \lambda_5 L_{VGG}
\end{aligned}
\end{equation}

Different from UNIT, MUNIT adds context code and style code reconstruction losses on top of $L_{total}$ defined for UNIT,
\begin{equation}
\begin{aligned}
    L_{MUNIT} = L_{total} + \lambda_c L_{recon{\text -}c} + \lambda_s L_{recon{\text -}s}
\end{aligned}
\end{equation}
where $\lambda_c$ and $\lambda_s$ are used to control the weight of these loss terms. See MUNIT \cite{MUNIT_ECCV_2018} for details of these losses.

\paragraph{Adversarial discriminative} We chose to implement a feature discriminator following ADA \cite{ADA_IROS_2017}. The feature discriminator aligns the marginal feature distributions of source and target domains hence increasing the performance of the detector with decision boundaries optimised on the source domain. The discriminator $D_{fea}$ is optimized jointly with the encoder $E$ through adversarial losses $L_{adv{\text -}fea}$:
\begin{equation}
\begin{aligned}
    L_{adv{\text -}fea}(D) = & \mathbb{E}_{X \sim p_{real}(X)}log(1-D(E(X))) + \\
    & \mathbb{E}_{X \sim p_{sim}(X)}log(D(E(X)))
\end{aligned}
\end{equation}
\begin{equation}
\begin{aligned}
    L_{adv{\text -}fea}(G) = & \mathbb{E}_{X \sim p_{real}(X)}log(D(E(X))) + \\
    & \mathbb{E}_{X \sim p_{sim}(X)}log(1-D(E(X)))
\end{aligned}
\end{equation}
Here the encoder is trained to maximize the domain confusion of the discriminator. In our experiments, we found that adding a feature discriminator also improves the performance of the adversarial generative method. Thus, the encoder-generator loss defined in equation \ref{total_loss} is modified to:
\begin{equation}
\begin{aligned}
    L_{total} = & \lambda_0 L_{recon} + \lambda_1 L_{cyc{\text -}recon} + \lambda_2 L_{LSGAN}(G) + \\
    & \lambda_3 L_{task} + \lambda_4 L_{cyc{\text -}task} + \lambda_5 L_{VGG} + \\
    & \lambda_6 L_{adv{\text -}fea}(G)
\end{aligned}
\end{equation}

\section{EXPERIMENT}

In this section, we evaluate our method on TuSimple, a widely used real-world lane detection dataset with classification labels of the training and validation set provided by \cite{Lane_Cls_2020}. As described in Section 3, the evaluated methods do not use labels in the real-world dataset during training, only real-world images and labelled simulation data generated with \emph{Simulanes}. By default, the number of images in the simulation dataset match the number of training images in the real-world dataset. In our ablation studies, we further explore the effect of simulation dataset size on performance. All experiments were run 3 times with different random seeds and the mean and standard deviation are reported. 

\subsection{Experimental Setup}

\paragraph{Dataset} TuSimple has 6,408 frames with 1280$\times$720 resolution which is split into 3,268 training, 358 validation, and 2,782 test images. TuSimple contains daytime highway scenarios in fair weather with varied traffic conditions. 

\paragraph{Evaluation metrics} For TuSimple, the official evaluation metric is accuracy, described by:
\begin{equation}
    Accuracy = \frac{\sum_{clip} C_{clip}}{\sum_{clip} S_{clip}}
\end{equation}
where $C_{clip}$ is the number of correctly predicted points in each frame of the clip, $S_{clip}$ is the number of ground truth points in each frame of the clip. A correctly predicted point is within a width threshold from the ground-truth. 

For classification, we follow the two classes used in \cite{Lane_Cls_2020}: dashed and continuous. Since TuSimple and CARLA had different lane classes available, we mapped each separately into the two classes. For TuSimple, we used the same mapping as \cite{Lane_Cls_2020} and for the additional CARLA classes, we mapped solid dashed lane combinations as continuous as well. 

\paragraph{Benchmarks} Our detection model is the UFLD model with a classification branch as described in Section III.B. The simplest method, direct transfer, is to train the detector on the synthetic dataset without any domain adaptation. Another method is to take a two-stage approach where UNIT or MUNIT is first trained to translate images from simulation to real-world, then the detector is trained on translated simulation dataset images. As an upper bound, we compare to the detector trained on the TuSimple dataset with labels.

\paragraph{Implementation details} For the lane model, we use the same gridding cell and row anchor configuration as \cite{UFLD_ECCV_2020} and the input image is resized to 288$\times$800. During training, data augmentation consisting of rotation, vertical and horizontal shift is added. The networks are optimized using ADAM \cite{ADAM_ICLR_2015} with a cosine decay learning rate schedule for a maximum of 100 epochs. The lane detector and discriminators use an initial learning rate of 0.0004 while the encoder-generator pairs use 0.0001. For the generative approach, we found that having a linear learning rate warmup from 0 to 0.0004 in 25 epochs was optimal and use $\lambda_0=\lambda_1=10$, $\lambda_2=\lambda_3=\lambda_4=\lambda_5=1$, $\lambda_6=0.1$ for the encoder-generator loss weights. For the lane detector, we use $\alpha=\beta=1$ and $\gamma=0.1$. Lastly, for MUNIT $\lambda_c=\lambda_s=1$. The validation set lane detection accuracy is used to select the best model for testing.

\subsection{Results}

Table \ref{results} summarizes our results on TuSimple. The detection accuracy was obtained using the test set, while the classification accuracy was only generated on the validation set as the test set does not have classification labels.
Compared to detection accuracy, the higher variance in classification scores could be due to the small validation set used to generate these results. During training, the models were saved based on the highest lane detection accuracy on the validation set, without considering the classification accuracy.

\begin{table}[t]
  \centering
  \small
    \caption{Comparison of our proposed methods and baselines on TuSimple lane detection and classification.}
    \begin{tabular}{lcccc} 
    \hline
      \multirow{2}{*}{Model} & \multicolumn{2}{c}{Det-Acc} & \multicolumn{2}{c}{Cls-Acc} \\
      
      & mean & stddev & mean & stddev \\
      \hline
      Direct transfer & 82.60 & 1.177 & 42.25 & 6.358 \\
      S2R translation (UNIT) & 77.55 & 4.485 & 34.19 & 0.772\\
      S2R translation (MUNIT) & 78.61 & 1.371 & 38.57 & 7.836\\
      Adv. discriminator (ADA) & \textbf{82.90} & 0.069 & 55.31 & 3.829\\
      MUNIT + adv. dis. & 82.28 & 1.605 & 53.70 &1.228\\
      UNIT + adv. dis. & 81.62 & 0.648 & \textbf{55.77} & 3.847\\
      \hline
      Detector trained on real & 95.38 & 0.130 & 37.08 & 0.185 \\
      \hline
    \end{tabular}
    \label{results}
\end{table}

Two-stage models consistently underperform other methods by 4$\sim$6\% in detection accuracy and classification accuracy suffers more, with both two-stage models averaging around 18\% worse. This poor performance is likely due to the translator not preserving the small lane features in the content of the image during the first training stage as the lane task losses are not included. Qualitatively, the translated images show very little contrast between road and lane markings. The lanes are difficult to identify in the translated image, meaning that the translation step makes the detection task more difficult for the downstream network.

MUNIT consistently outperforms UNIT in detection and classification in the proposed method and in image translation. This can be a result of separating the context and style codes in the latent space allowing MUNIT to have a slightly larger feature space to store context information for the detector.
Our adversarial discriminative method (ADA) outperforms the adversarial generative methods. This may be due to the small size of TuSimple, making it more difficult for generative methods to converge. Thus, ADA only optimizing for feature similarity via the discriminator was an easier task.

\begin{figure}[t]
\centering
\includegraphics[width=0.95\columnwidth]{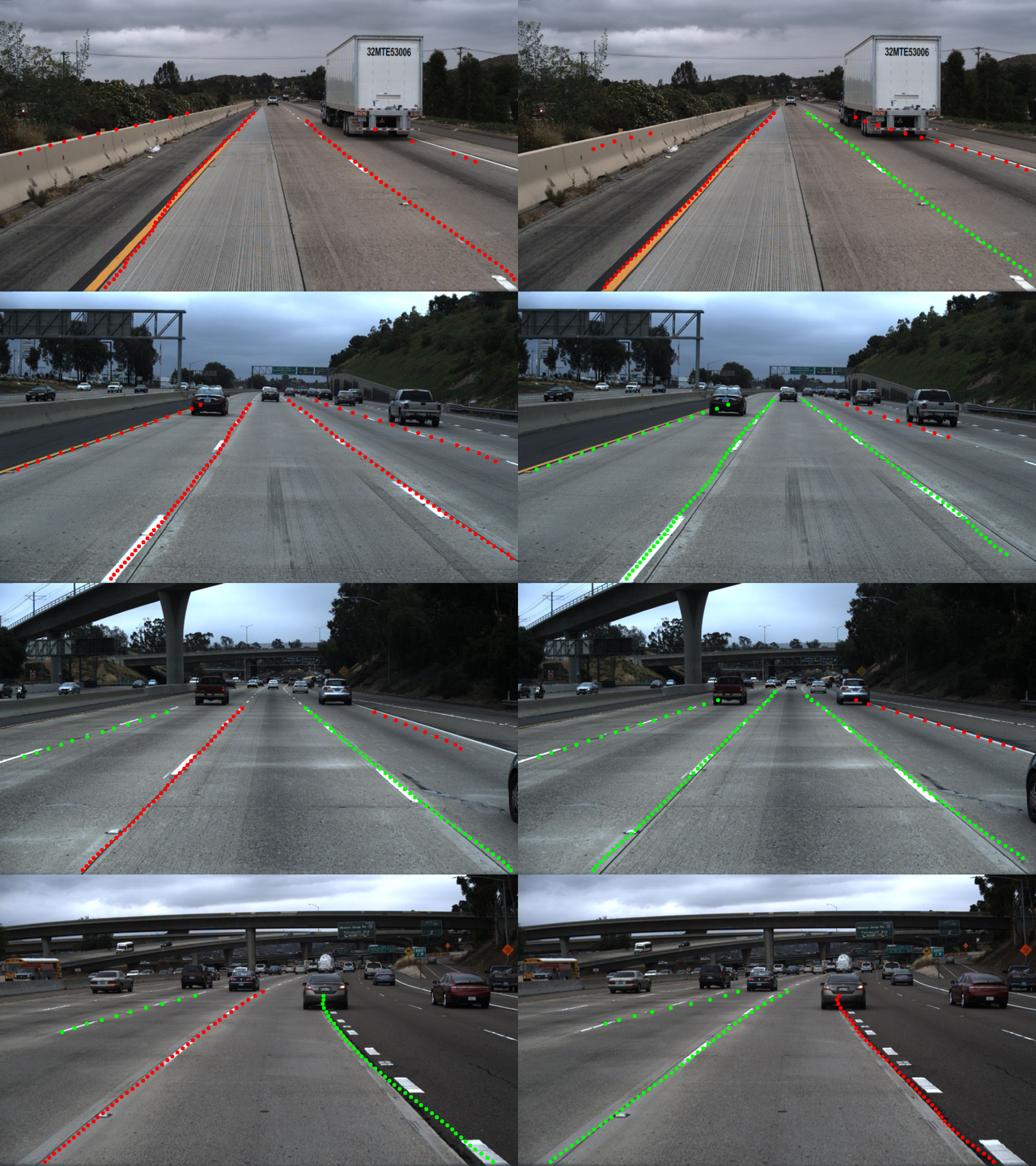} 
\caption{Sample images from TuSimple test set. Predictions from direct transfer method on the left column and ADA on the right. The lane classes are labelled using red and green for solid and dashed respectively.}
\label{lane_viz}
\end{figure}

We observe that the proposed methods perform similarly in detection but outperform direct transfer method by a margin in classification. This suggests that our simulation dataset is close to TuSimple in terms of lane location but more different in lane appearance, creating a bigger domain gap for classification. This is further shown in Figure \ref{lane_viz}. 

Lastly, the results presented can be a starting point for future works using our \emph{Simulanes} dataset. 
There still exists a gap in detection performance of domain adaptation methods in comparison to the fully supervised upper bound, whereas lane classification accuracy can be improved overall.
Semi-supervised methods, as a middle ground cost-wise between fully labelled and unsupervised approaches, can also be explored and evaluated using \emph{Simulanes}. 

\subsection{Ablation Study}

\begin{figure}[h]
\centering
\includegraphics[width=1.0\columnwidth]{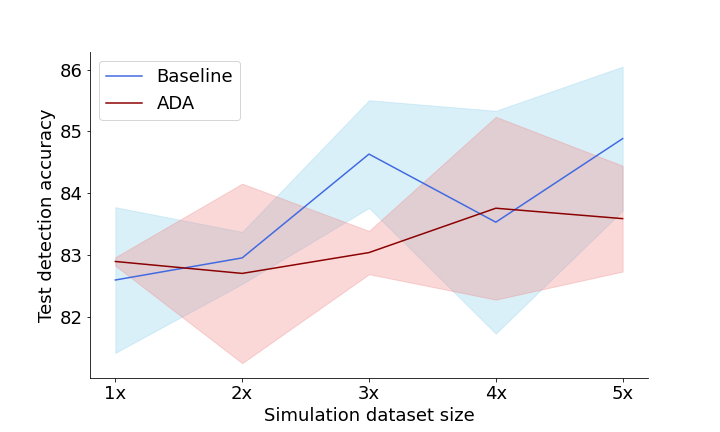} 
\caption{Impact of the simulated dataset size on test set detection accuracy.}
\label{ablation_det_plot}
\end{figure}

\begin{figure}[h]
\centering
\includegraphics[width=1.0\columnwidth]{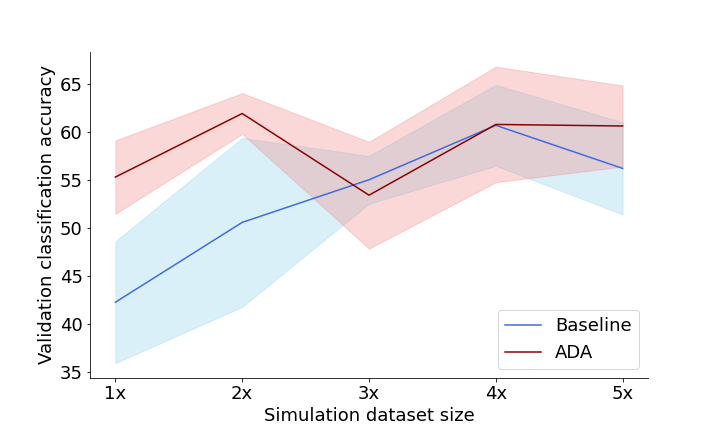} 
\caption{Impact of the simulated dataset size on validation set classification accuracy.}
\label{ablation_cls_plot}
\end{figure}

For our ablation study, we examine the effect of the simulation dataset size on the test set performance. As generating more images for the simulation dataset is more feasible compared to collecting and labelling real-world images, seeing an increase in test set performance by simply increasing the simulation dataset size would be beneficial. Since the adversarial discriminative approach was found to be our best performing method, we ran this study with ADA and the direct transfer baseline which is trained only using labelled simulation data. For the study, we generated a simulation dataset 5x the size of the TuSimple training set, resulting in a dataset of 16344 images. We then trained the baseline and ADA on 5 subsets of this dataset, each with the number of images being an integer multiple of the TuSimple training set size, ranging from 1x to 5x. Considering ADA needs a real image and a simulated image for each training step, we trained using a 1x number of simulation images at each epoch but re-sampled between epochs for both models. The results for the detection and classification accuracy are shown in Figures \ref{ablation_det_plot} and \ref{ablation_cls_plot}, respectively. The dark line represents the mean and the shaded area shows the standard deviation from the mean in both directions. We observe a general increase in both detection and classification accuracy as the simulation dataset size increases. 

\section{CONCLUSIONS}

In this paper, we propose unsupervised domain adaptation techniques using adversarial generative and feature discriminate approaches for lane detection and classification applications in autonomous driving. To facilitate UDA, we introduced \emph{Simulanes}, a simulation data generator for lane detection and classification using CARLA. Using our synthetic dataset and TuSimple, we evaluated our proposed method against benchmarks in unsupervised sim-to-real domain adaptation. We observed that with a fixed number of real-world images, we can improve detection and classification performance by increasing the number of synthetic images available to the model.
The \emph{Simulanes} dataset generation tool can be used in future methods for evaluating works in sim-to-real domain adaptation and producing lane detection models without the use of real-world labelled data. In future work, we aim to extend the learnt UDA-based models to more complicated scenarios, e.g. night images, presented in \cite{CULane_Dataset}.

\addtolength{\textheight}{-1cm}   


\section*{ACKNOWLEDGMENT}

This work was supported by NSERC CRD 537104-18, in partnership with General Motors Canada and the SAE AutoDrive Challenge.


{\small
\bibliographystyle{IEEEtran}
\bibliography{IEEEabrv, references}
}

\end{document}